\documentclass{article}
\usepackage{arxiv}
\usepackage[numbers]{natbib}
\usepackage[utf8]{inputenc} % allow utf-8 input
\usepackage[T1]{fontenc}    % use 8-bit T1 fonts
\usepackage[hidelinks]{hyperref} % hyperlinks (with hidden links)
\usepackage{url}            % simple URL typesetting
\usepackage{booktabs}       % professional-quality tables
\usepackage{amsfonts}       % blackboard math symbols
\usepackage{nicefrac}       % compact symbols for 1/2, etc.
\usepackage{microtype}      % microtypography
\usepackage{graphicx}       % for including graphics
\usepackage{natbib}         % for citations
\usepackage{doi}            % for DOI formatting
\usepackage{enumitem}       % for custom lists
\usepackage[ruled,vlined]{algorithm2e} % for algorithm formatting

\title{Perturbations in the Orthogonal Complement Subspace for Efficient Out-of-Distribution Detection}

\author{
  Zhexiao Huang\\
  School of Mathematics and Statistics\\
  Guangdong University of Technology\\
  Guangzhou, China\\
  \and
  Weihao He\\
  School of Mathematics and Statistics\\
  Guangdong University of Technology\\
  Guangzhou, China\\
  \and
  Shutao Deng\\
  School of Mathematics and Statistics\\
  Guangdong University of Technology\\
  Guangzhou, China\\
  \and
  Junzhe Chen\\
  School of Mathematics and Statistics\\
  Guangdong University of Technology\\
  Guangzhou, China\\
  \and
  Chao Yuan\\
  School of Mathematics and Statistics\\
  Guangdong University of Technology\\
  Guangzhou, China\\
  \and
  Hongxin Wang\\
  School of Mathematics and Statistics\\
  Guangdong University of Technology\\
  Guangzhou, China\\
  \and
  Changsheng Zhou*\\
  School of Mathematics and Statistics\\
  Guangdong University of Technology\\
  Guangzhou, China\\
  \href{mailto:chsh_zh@gdut.edu.cn}{\nolinkurl{chsh_zh@gdut.edu.cn}}
}

\renewcommand{\shorttitle}{\textit{arXiv} Template}

\hypersetup{
pdftitle={Perturbations in the Orthogonal Complement Subspace for Efficient Out-of-Distribution Detection},
pdfsubject={cs.LG, cs.CV, stat.ML},
pdfauthor={Zhexiao~Huang and Weihao He and Shutao Deng and Junzhe Chen and Chao Yuan and Hongxin Wang and Changsheng~Zhou},
pdfkeywords={out-of-distribution detection, OOD, orthogonal complement subspace, perturbation, robustness, deep learning},
}

\begin{document}
\maketitle

\begin{abstract}
	Out-of-distribution (OOD) detection is indispensable for the reliable deployment of deep learning models in open-world environments. Existing approaches, including energy-based scoring and gradient-projection methods, typically exploit high-dimensional representations to separate in-distribution (ID) from OOD samples. We present P-OCS (Perturbations in the Orthogonal Complement Subspace), a lightweight and theoretically grounded method that operates within the orthogonal complement of the principal subspace spanned by ID features. P-OCS applies a single projected perturbation confined to this complementary subspace, selectively amplifying subtle ID–OOD discrepancies while preserving the geometry of ID representations. We show that, in the small-perturbation limit, a one-step update is sufficient and provide convergence guarantees for the resulting detection score. Extensive experiments across diverse architectures and datasets demonstrate that P-OCS achieves state-of-the-art OOD detection with negligible computational overhead, without requiring model retraining, access to OOD data, or architectural modifications.
\end{abstract}


\section{Introduction}

Deep neural networks frequently produce highly confident predictions when confronted with samples drawn from distributions beyond their training regime. 
As such, identifying out-of-distribution (OOD) samples is essential for the safe deployment of models in open-world settings. 
A substantial body of prior work has explored post-hoc scoring mechanisms—such as ODIN~\cite{Liang2018ODIN}, the Mahalanobis-distance method~\cite{Lee2018Mahalanobis}, energy-based scores~\cite{Liu2020EnergyOOD}, and GradOrth~\cite{Behpour2023GradOrth}—to distinguish in-distribution (ID) from OOD samples using features of pretrained classifiers. 
These methods tend to perform well when OOD data exhibit substantial covariate or feature shift (i.e., \emph{far-OOD}), but their performance often degrades in more challenging regimes of \emph{semantic shift with feature overlap} (i.e., \emph{near-OOD}), even when aided by post-processing strategies such as ReAct or PCA-based removal of dominant principal components.

In our investigation, we observe a consistent geometric phenomenon in the \emph{penultimate-layer feature space}: 
ID samples concentrate variance within a dominant principal subspace, whereas OOD samples distribute variance more broadly into the orthogonal complement of that subspace. 
This observation motivates our central insight: 
\emph{a single perturbation restricted to the orthogonal complement can expose intrinsic separability between ID and OOD samples, including in near-OOD regimes with strong feature overlap}.

Accordingly, we propose \textbf{P-OCS} (\textbf{P}erturbations in the \textbf{O}rthogonal \textbf{C}omplement \textbf{S}ubspace) — a minimalist yet theoretically grounded framework that performs a one-step orthogonal perturbation in the complement of the ID principal subspace, computed at the penultimate layer. 
Empirically, this single iteration achieves near-optimal discriminative power with negligible computational overhead.

\begin{figure}[ht]
  \centering
  \begin{minipage}[b]{0.24\linewidth}
    \centering
    \includegraphics[width=\textwidth,height=0.63\textwidth]{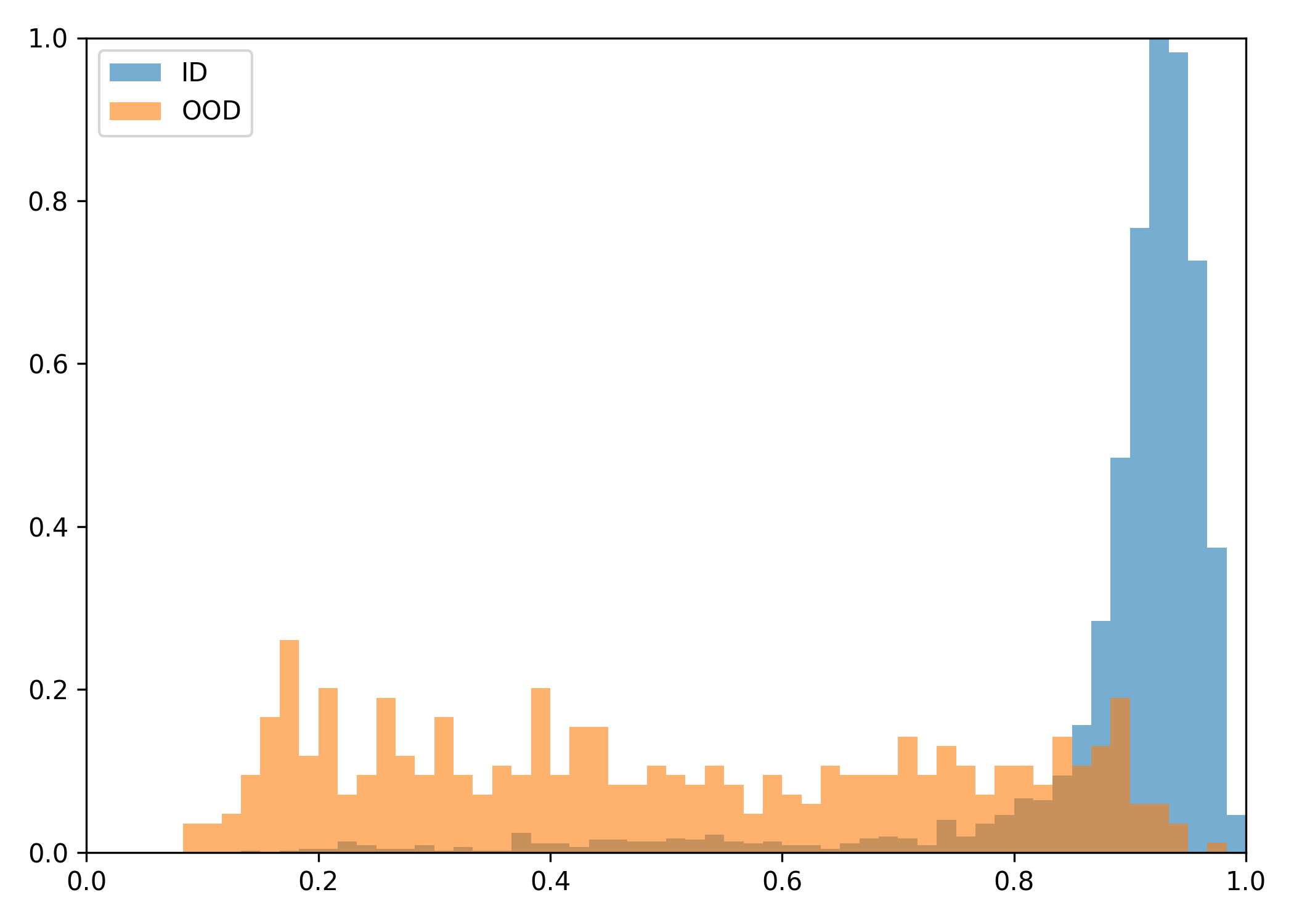}
    \caption*{(a) ReAct + MSP}
  \end{minipage}
  \hfill
  \begin{minipage}[b]{0.24\linewidth}
    \centering
    \includegraphics[width=\textwidth]{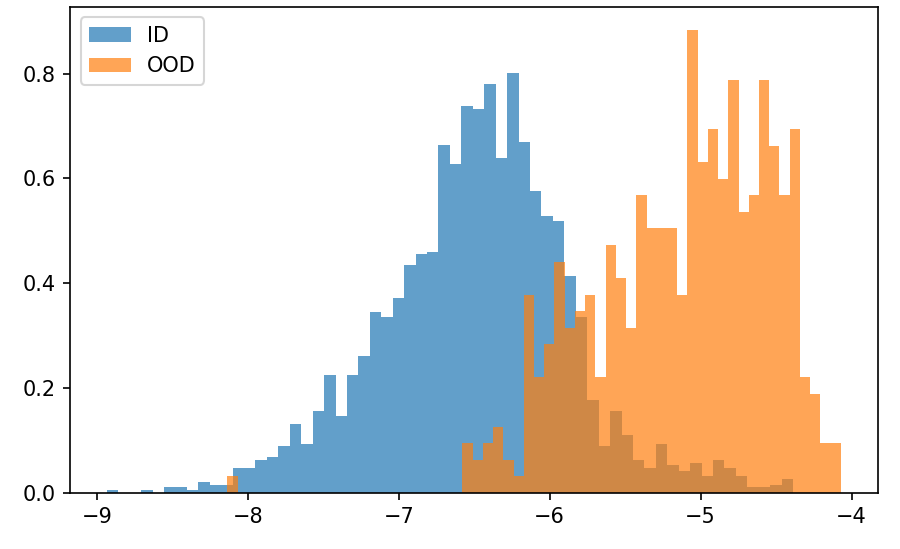}
    \caption*{(b) ReAct + Energy}
  \end{minipage}
  \hfill
  \begin{minipage}[b]{0.24\linewidth}
    \centering
    \includegraphics[width=\textwidth]{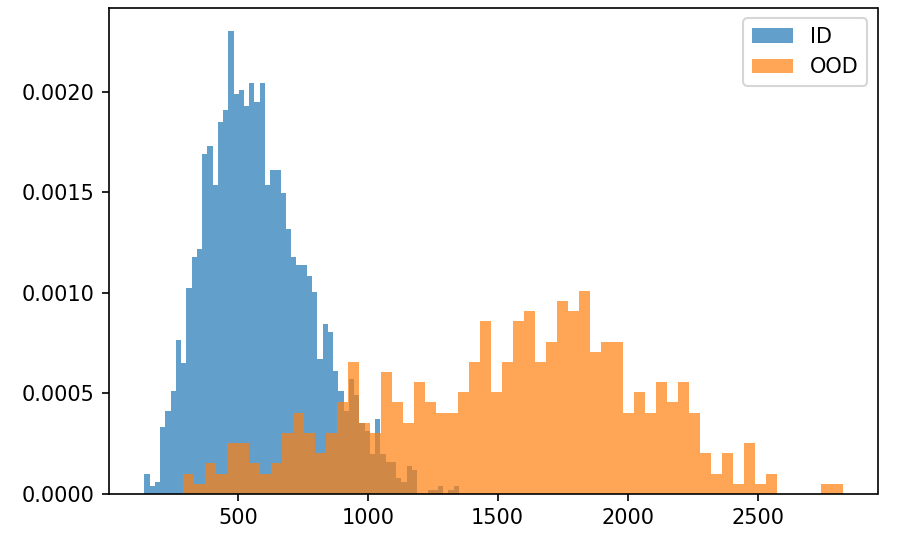}
    \caption*{(c) ReAct + Mahalanobis}
  \end{minipage}
  \hfill
  \begin{minipage}[b]{0.24\linewidth}
    \centering
    \includegraphics[width=\textwidth,height=0.63\textwidth]{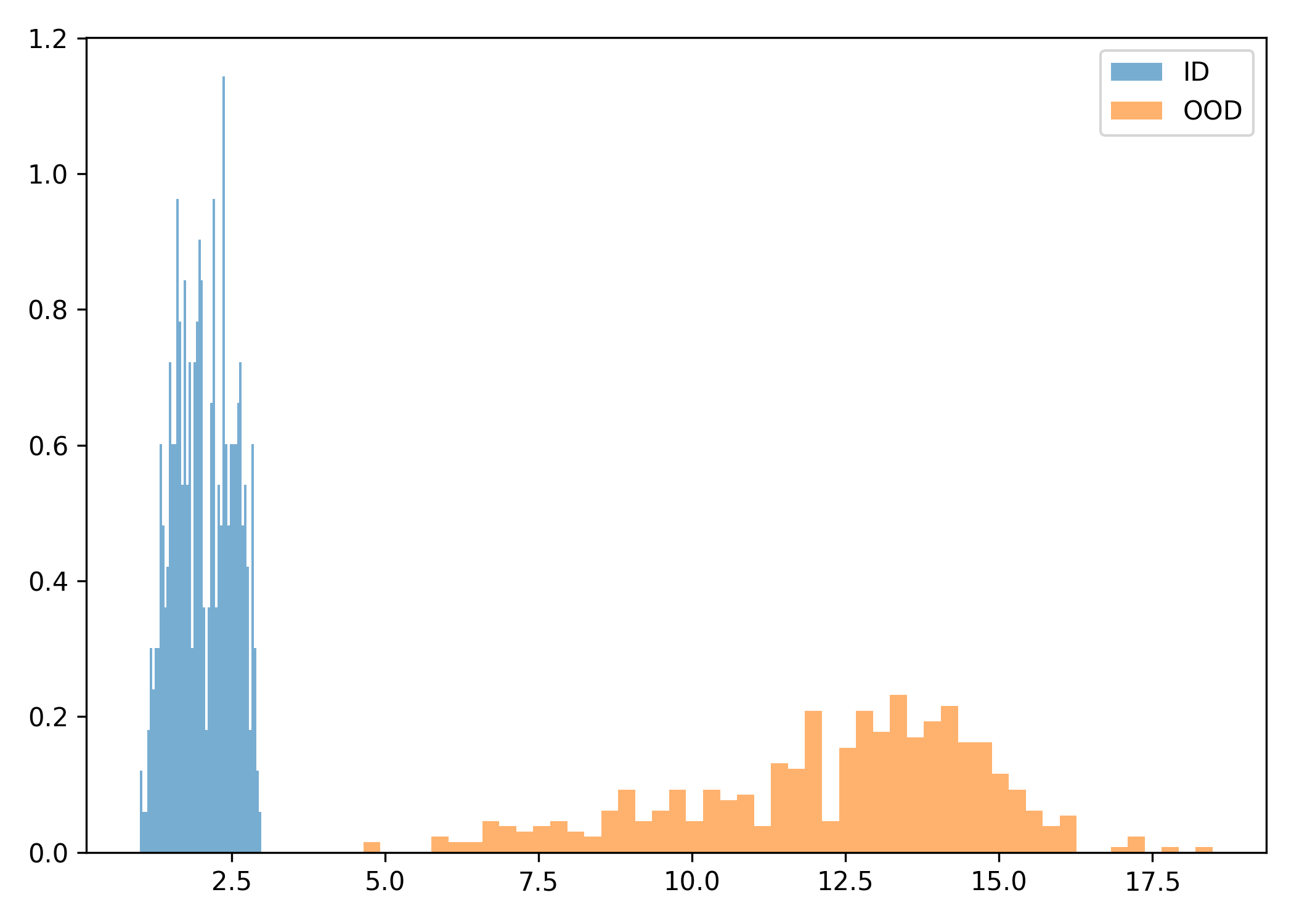}
    \caption*{(d) P-OCS}
  \end{minipage}
  \caption{
    Comparison of score distributions for ID (green) and Near-OOD (blue) samples under different post-hoc scoring schemes.
    Existing methods (a–c) show significant overlap between ID and OOD scores, indicating poor separability.
    In contrast, our proposed P-OCS (d) yields a clear margin between the two distributions with minimal computation.
  }
  \label{fig:score_dists}
\end{figure}

To better visualize the difference between traditional post-hoc scores and our proposed P-OCS, we compare the score distributions of ID and OOD samples under four representative scoring schemes: (a) ReAct-processed Maximum Softmax Probability (MSP), (b) ReAct-processed Energy score, (c) ReAct-processed Mahalanobis distance, and (d) our proposed P-OCS score. 
As shown in Fig.~\ref{fig:score_dists}, existing methods exhibit substantial overlap between ID and OOD distributions in near-OOD regimes, while P-OCS yields a distinct and well-separated score boundary.

\begin{figure}[ht]
  \centering
  \begin{minipage}[b]{0.4\linewidth}
    \centering
    \includegraphics[width=\textwidth,height=0.83\textwidth]{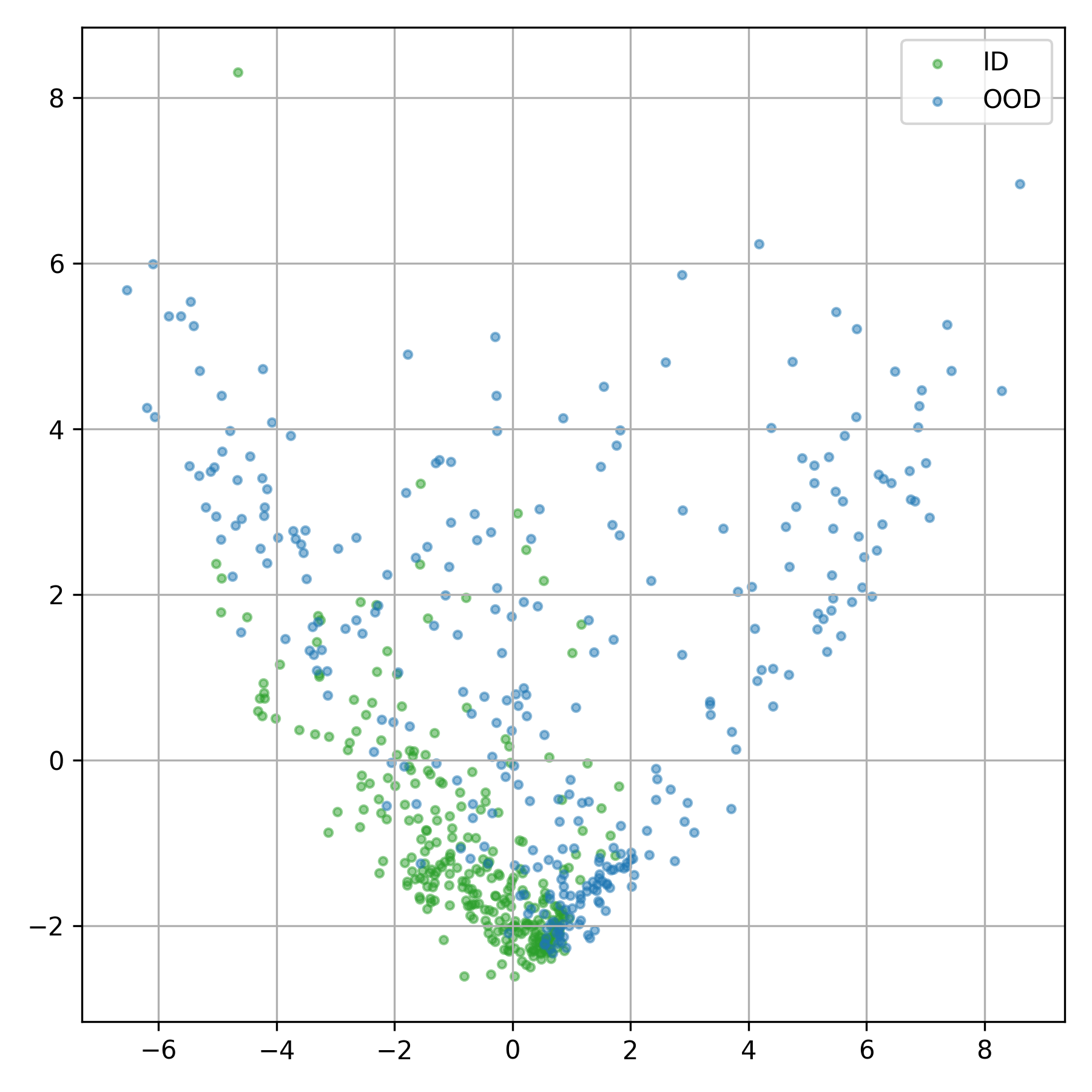}
    \caption{PCA projection of ID and Near-OOD samples. ID samples (green) concentrate along the principal components, while Near-OOD samples (blue) exhibit greater dispersion across orthogonal directions. This illustrates the challenge in separating Near-OOD samples from ID using standard methods.}
    \label{fig:pca1}
  \end{minipage}
  \hfill
  \begin{minipage}[b]{0.4\linewidth}
    \centering
    \includegraphics[width=\textwidth,height=0.83\textwidth]{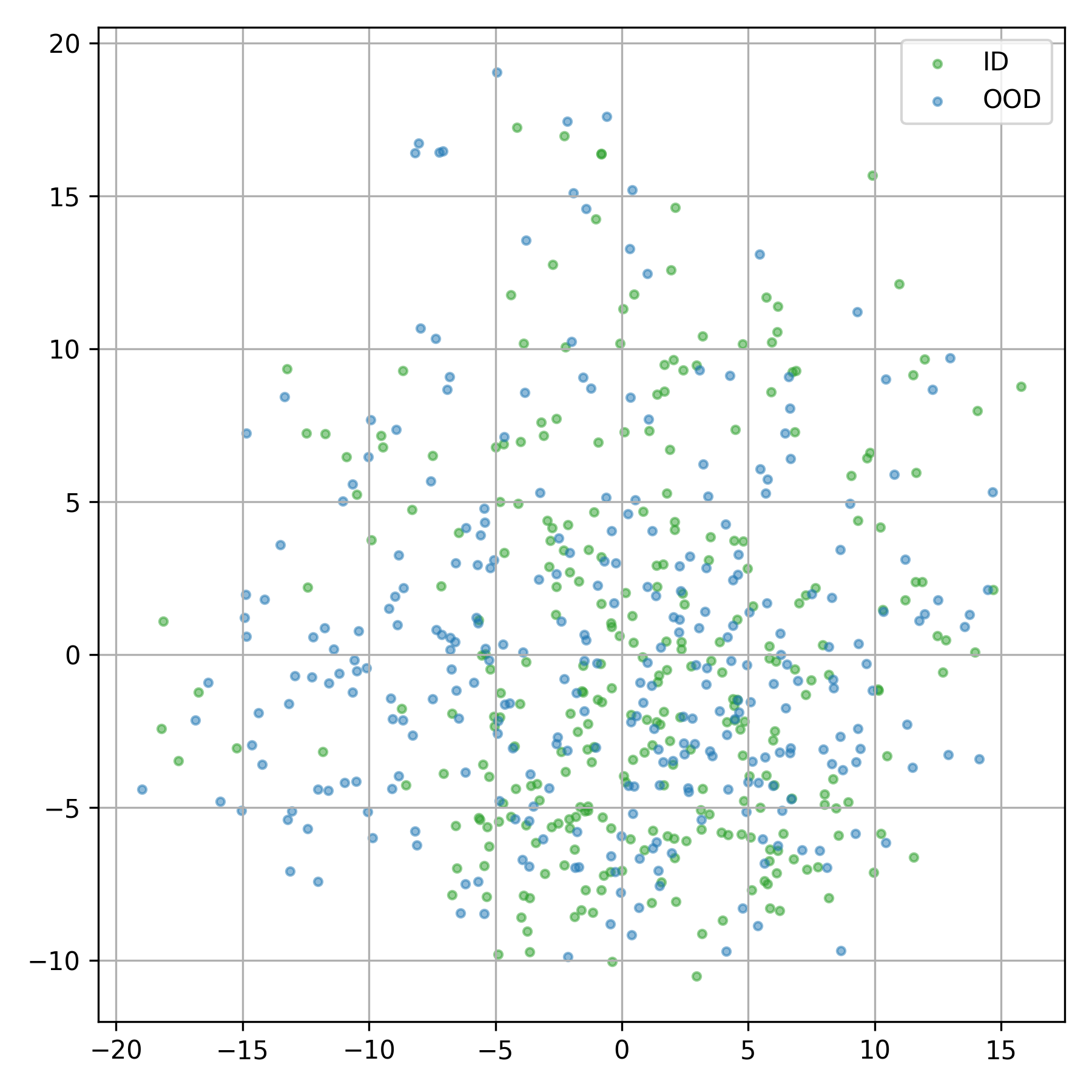}
    \caption{Further PCA analysis highlighting the challenge of distinguishing Near-OOD from ID samples. The right plot shows how Near-OOD samples (blue) are dispersed across the feature space, making it difficult for traditional models to separate them from ID (green) samples without additional processing.}
    \label{fig:pca2}
  \end{minipage}
\end{figure}

\textbf{Contributions.} 
\begin{enumerate}[left=0pt, noitemsep]
  \item We introduce \textbf{P-OCS}, a one-step orthogonal-complement perturbation method for OOD detection that operates on penultimate-layer features, requiring no additional training or architectural modification.
  \item We provide theoretical justification that a single perturbation step suffices under small-perturbation regimes, yielding a principled detection score.
  \item We analyze the variance structure of the orthogonal complement and establish its connection to distributional separability, clarifying why near-OOD cases benefit markedly.
  \item We demonstrate state-of-the-art detection performance and computational efficiency across multiple benchmarks and architectures, with consistent gains on \emph{near-OOD} as well as strong results on \emph{far-OOD}.
\end{enumerate}

\section{Related Work}

\subsection{Out-of-Distribution Detection}
Classical OOD detection approaches primarily rely on post-hoc confidence estimation over pretrained classifiers. Early methods include the maximum softmax probability (MSP) baseline~\cite{Hendrycks2017Baseline}, temperature scaling and input perturbation in ODIN~\cite{Liang2018ODIN}, Mahalanobis distance-based detectors~\cite{Lee2018Mahalanobis}, and energy-based scoring~\cite{Liu2020EnergyOOD}. 
Recent advances, such as ReAct~\cite{Sun2021ReAct}, DICE~\cite{Sun2022DICE}, and GradOrth~\cite{Behpour2023GradOrth}, enhance robustness by modifying activation statistics or gradient representations. 
While these techniques improve detection in far-OOD settings, many remain computationally expensive or lack geometric interpretability, especially for semantically shifted near-OOD cases.

\subsubsection{Subspace and Spectral Methods}
Principal component analysis (PCA) and singular value decomposition (SVD) have been widely used to characterize the intrinsic structure of in-distribution (ID) features~\cite{Ren2019Likelihood, Tack2020CSI}. 
For instance, GradOrth~\cite{Behpour2023GradOrth} projects gradients onto dominant singular directions to suppress irrelevant variance, aligning feature responses along stable axes. 
Our work extends this perspective by explicitly leveraging the \emph{orthogonal complement subspace}, which captures variance components where OOD deviations predominantly reside. 
This formulation bridges subspace geometry with probabilistic separability in OOD detection.

\subsubsection{Perturbation and Feature Stability}
Perturbation-based and adversarial methods have long studied the stability of model predictions under small input or feature perturbations~\cite{Goodfellow2015AdvTraining, Hein2019ReluSpheres}. 
Unlike input-space perturbations, our approach introduces a iterative perturbation directly in the learned orthogonal feature manifold. 
This design leads to a mathematically interpretable dynamic process that exposes intrinsic ID–OOD separability without requiring adversarial optimization or retraining.

\section{Method}

\subsection{Preliminaries}
Let $ X_{\mathrm{ID}} \in \mathbb{R}^{N \times d} $ denote the features extracted from an in-distribution (ID) dataset. 
We first compute its mean and perform principal component analysis (PCA)~\cite{Jolliffe2016PCA,Bishop2006Pattern}:
\begin{equation}
    X_c = X_{\mathrm{ID}} - \mu, \quad X_c = U \Sigma V^{\top},
\end{equation}
where $ U = [U_k, U_\perp] $ separates the feature space into the principal subspace $ U_k $ and its orthogonal complement $ U_\perp $.  
We define the projection operators:
\begin{equation}
    P = U_k U_k^\top, \quad P_\perp = U_\perp U_\perp^\top.
\end{equation}

To visualize the variance ratio explained by the ID and OOD projections onto the ID basis and the complement space, we present the following two figures. These figures show how the variance is distributed across the first 80 components of the ID and complement spaces, highlighting the differences between the ID and OOD distributions.

\begin{figure}[ht]
    \centering
    \begin{minipage}{0.48\textwidth}
        \centering
        \includegraphics[width=\textwidth]{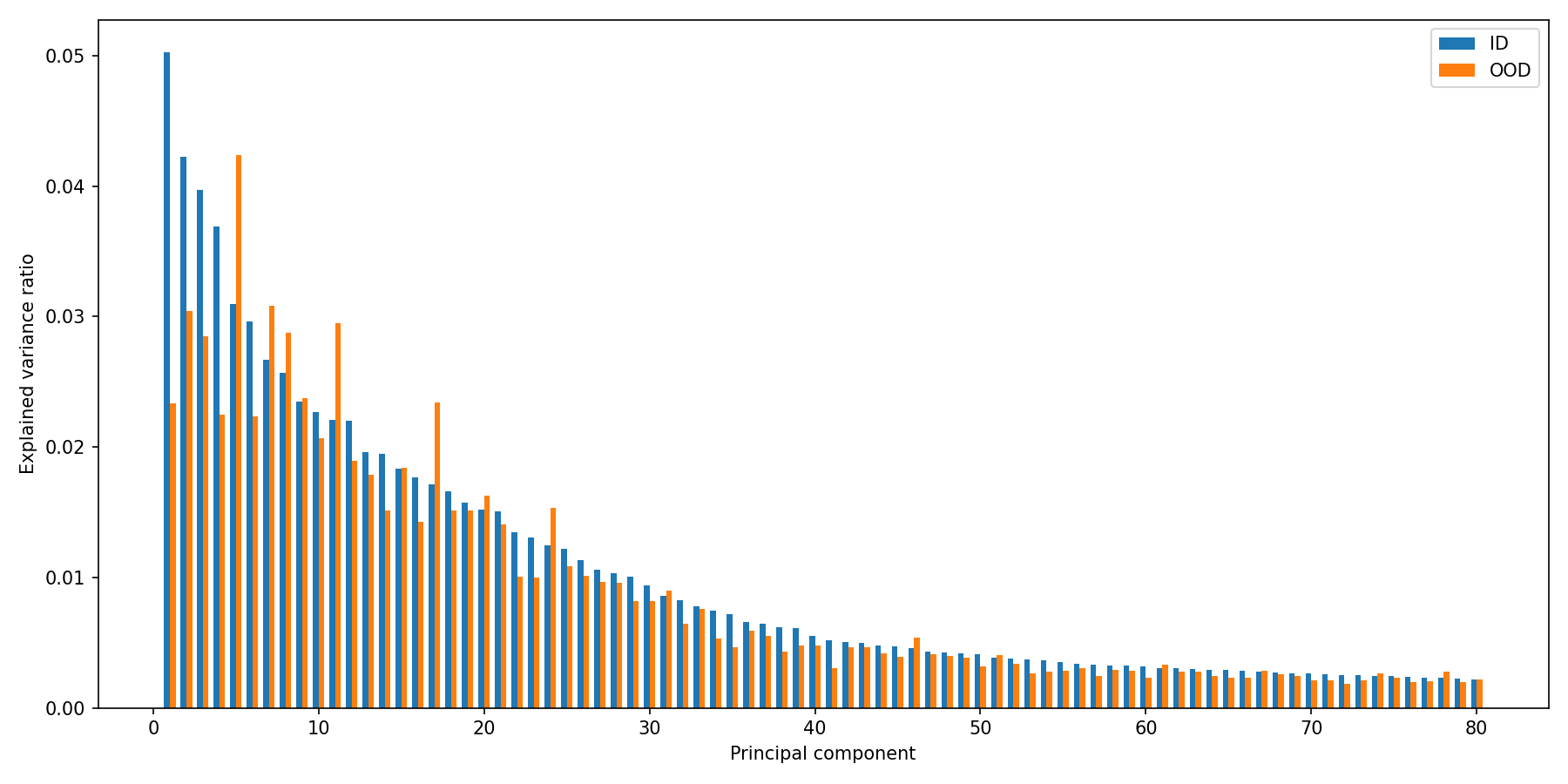}
        \caption{ID vs OOD explained variance ratio (first 80 components) — on ID basis.}
        \label{fig:id_vs_ood_id_basis}
    \end{minipage}
    \hfill
    \begin{minipage}{0.48\textwidth}
        \centering
        \includegraphics[width=\textwidth]{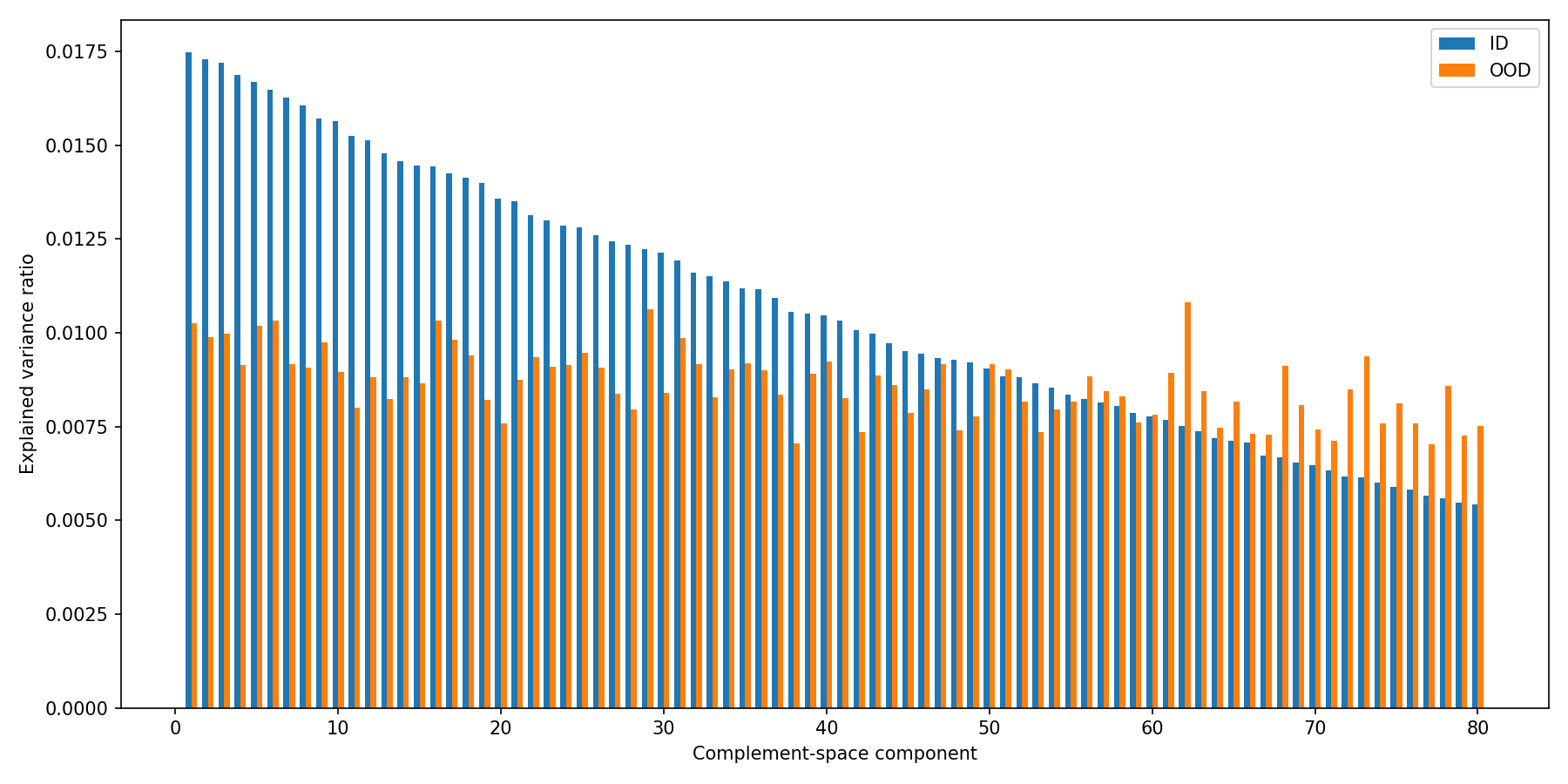}
        \caption{ID vs OOD explained variance ratio in complement space (first 80 components).}
        \label{fig:id_vs_ood_complement_space}
    \end{minipage}
\end{figure}

\textbf{Explanation of Figures:}  
The first figure (\ref{fig:id_vs_ood_id_basis}) presents the explained variance ratio for the first 80 principal components of the ID space. As expected, the ID distribution shows a higher variance in the first few components. The second figure (\ref{fig:id_vs_ood_complement_space}) presents the explained variance ratio in the complement space, showing how the OOD samples behave when projected onto the orthogonal subspace $U_\perp$. This comparison reveals how OOD samples tend to occupy regions of the feature space that explain lower variance, consistent with their dissimilarity to the ID samples~\cite{yang2023geometric}.

\subsection{Orthogonal Complement Perturbation}
We define a stochastic perturbation matrix in the orthogonal complement space:
\begin{equation}
    A = [(1 - \varepsilon)I + \varepsilon Q_{\mathrm{orth}}] D,
\end{equation}
where $ Q_{\mathrm{orth}} $ is a random orthogonal matrix and $ D $ is a random diagonal scaling matrix~\cite{mcallester2023orthogonal}.  
The perturbation intensity is controlled by $\varepsilon \in [0,1]$.

Given an input feature $ z_0 $, we iteratively update it for $T$ steps as:
\begin{equation}
    z_{t+1} = P z_t + U_\perp A_t U_\perp^\top z_t, \quad t = 0, 1, \dots, T-1,
\end{equation}
where each $A_t$ is independently sampled at every step.  
This iterative propagation models the feature dynamics in the orthogonal complement space, allowing the method to capture subtle deviations induced by distributional shifts.

\subsection{OOD Score Based on Accumulated Step Length}
To measure the instability of the feature under orthogonal perturbations, we define the OOD score as the total accumulated displacement of $z$ across all iterations:
\begin{equation}
    s(x) = \sum_{t=0}^{T-1} \| z_{t+1} - z_t \|_2.
\end{equation}
This formulation captures the overall response of the representation to orthogonal perturbations — a more OOD-like sample tends to exhibit larger cumulative drift.  
The metric unifies variance-based and energy-based OOD principles~\cite{Liu2020EnergyOOD,Sun2021ReAct,Lee2018Mahalanobis} under a geometric framework.

\subsection{Algorithm for OOD Detection}
We summarize the proposed OOD detection procedure based on orthogonal complement dynamics (OCD) in Algorithm~\ref{alg:ocd}.  
Given a pretrained feature extractor $f(\cdot)$, the PCA basis of in-distribution (ID) features $(U_k, U_{\perp})$, and a test sample $x$, the algorithm estimates the OOD score $s(x)$ by measuring feature displacement under orthogonal complement perturbations.

\begin{algorithm}[htbp]
\caption{Orthogonal Complement Dynamics (OCD) for OOD Detection}
\label{alg:ocd}
\SetAlgoLined
\KwIn{Feature extractor $f(\cdot)$; ID PCA basis $(U_k, U_{\perp})$; test sample $x$; iteration number $T$; perturbation strength $\varepsilon$}
\KwOut{OOD score $s(x)$}

\BlankLine
\textbf{Initialization:} Compute centered feature $z_0 \leftarrow f(x) - \mu$\;
$s(x) \leftarrow 0$\;

\BlankLine
\For{$t \leftarrow 0$ \textbf{to} $T-1$}{
    Sample a random orthogonal perturbation matrix: $A_t \leftarrow [(1-\varepsilon)I + \varepsilon Q_t] D_t$\;
    Update the feature representation: $z_{t+1} \leftarrow Pz_t + U_{\perp} A_t U_{\perp}^\top z_t$\;
    Accumulate displacement: $s(x) \leftarrow s(x) + \|z_{t+1} - z_t\|_2$\;
}
\Return{$s(x)$}\;

\end{algorithm}

The computational complexity per sample is $\mathcal{O}(T \cdot d^2)$, dominated by matrix multiplications in the orthogonal complement space.
This algorithm estimates the geometric sensitivity of a feature representation to perturbations within the orthogonal complement subspace.  
A larger accumulated displacement $s(x)$ indicates stronger instability of the feature under orthogonal complement dynamics, implying a higher likelihood that the sample $x$ originates from an out-of-distribution (OOD) region.

\section{Experiment}

\subsection{Experimental Setup}
We conduct extensive experiments to evaluate the proposed P-OCS (Principal-Orthogonal Complement Score) method on both specialized and general-purpose visual datasets.
All experiments are implemented using PyTorch and executed on NVIDIA RTX1650.

\paragraph{Datasets.}
We first evaluate our approach on a dermatological image dataset containing diverse skin disease categories. This dataset serves as a challenging testbed for out-of-distribution (OOD) detection under high intra-class variability.
To further validate the generalization capability, we adopt the ImageNet validation set as the in-distribution (ID) data and employ three widely used OOD benchmarks: SUN397~\cite{Xiao2010SUN}, iNaturalist 2021~\cite{iNaturalist2021}, and DTD~\cite{Cimpoi2014DTD}.
These datasets respectively represent scene images, fine-grained natural categories, and texture patterns, thus covering a broad spectrum of distributional shifts.

\paragraph{Baselines.}
We compare P-OCS with several representative OOD detection methods built upon feature-level rectification, including:
\begin{itemize}[leftmargin=1.5em]
\item \textbf{ReAct + Energy}~\cite{Sun2021ReAct, Liu2020EnergyOOD}: a feature rectification method combined with energy-based scoring.

\item \textbf{ReAct + Mahalanobis}~\cite{Sun2021ReAct, Lee2018Mahalanobis}: replacing the energy score with a Mahalanobis distance metric.

\item \textbf{ReAct + MSP}~\cite{Sun2021ReAct,Hendrycks2017Baseline}: a variant using the maximum softmax probability as the OOD score.

\end{itemize}

All baselines are reimplemented under the same backbone for fair comparison.

\paragraph{Backbones.}
We evaluate our method using two representative architectures: ResNet-50~\cite{He2016ResNet} and ConvNeXt~\cite{Liu2022ConvNeXt}, both pretrained on ImageNet-1K.
For the dermatological dataset, ConvNeXt is used as the primary backbone due to its superior feature representation for fine-grained visual tasks.

\subsection{Results on Dermatological Dataset}

\begin{figure}[htbp]
\centering
\includegraphics[width=0.75\linewidth]{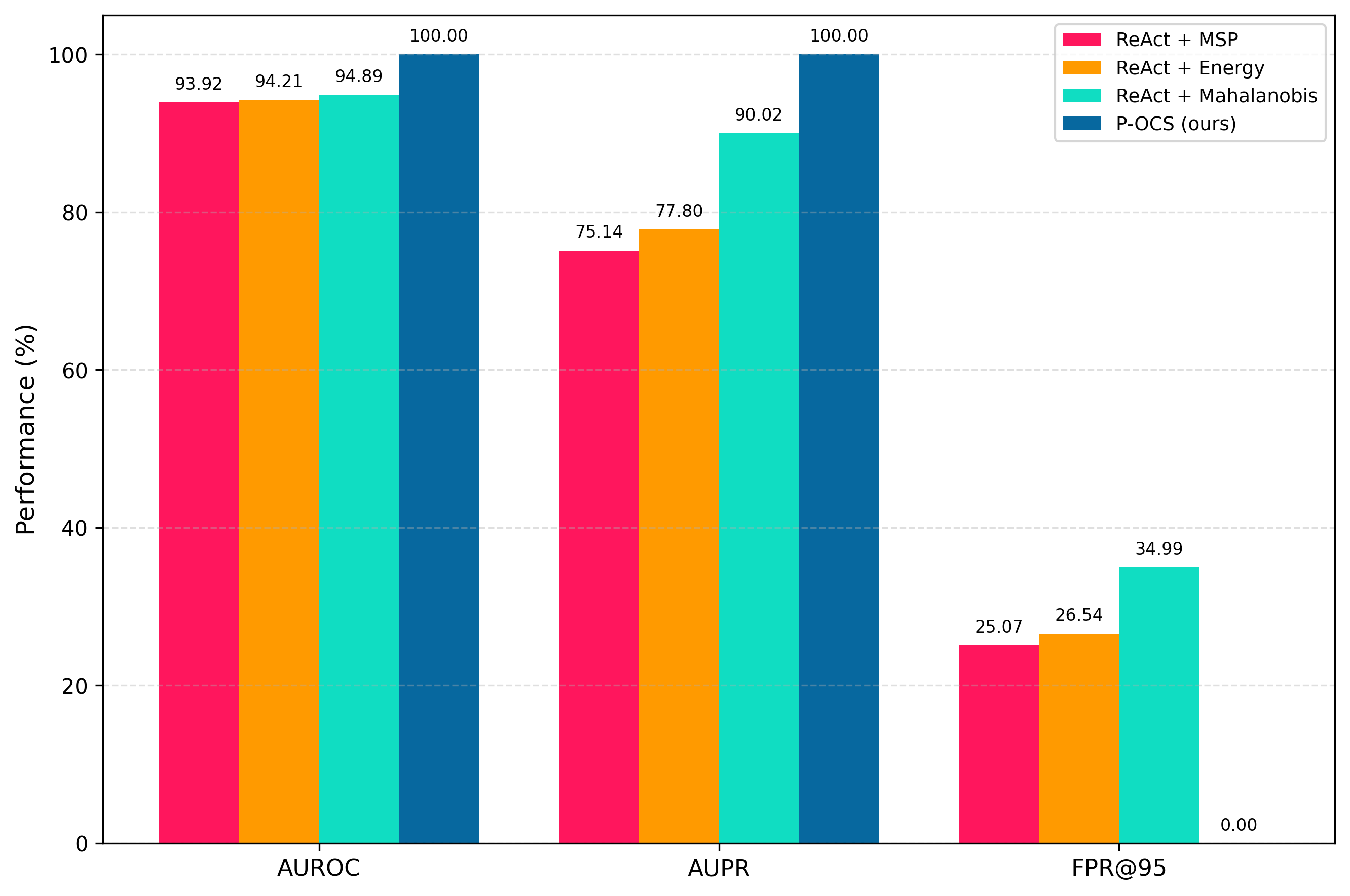}
\caption{Dermatology results with a metric-wise layout (x-axis: AUROC, AUPR, FPR@95). 
P-OCS consistently leads across all metrics, combining stronger discrimination and precision–recall performance with substantially reduced high-recall false positives.}
\label{fig:derma}
\end{figure}

\figurename~\ref{fig:derma} adopts a metric-wise layout (x-axis: \emph{AUROC}, \emph{AUPR}, \emph{FPR@95}), grouping competing methods within each metric. 
Across all three criteria, \textbf{P-OCS} exhibits a clear and uniform lead: it delivers stronger discrimination (AUROC), superior precision–recall behavior (AUPR), and markedly lower high-recall false positives (FPR@95) relative to rectification- and energy-based baselines. 
The improvements are simultaneous rather than metric-specific, indicating an overall lift in detection quality rather than a trade-off among objectives. 
Notably, the dermatology dataset features substantial visual similarity among classes, under which P-OCS maintains a consistent advantage, reflecting robustness under challenging overlap.

Methodologically, P-OCS explicitly targets \emph{orthogonal perturbation dynamics} in the feature space: the principal subspace is estimated on in-distribution (ID) data, and OOD scores are derived from responses in its orthogonal complement. 
This construction isolates OOD-sensitive directions that conventional scoring functions tend to under-emphasize, yielding a more reliable geometric separation between ID and OOD representations.

\paragraph{Protocol.}
All methods are evaluated under the same backbone and preprocessing. 
Hyperparameters are chosen on ID validation data only; OOD data are not used for model selection. 
We report AUROC, AUPR, and FPR@95 following standard practice on identical splits for all methods.

\subsection{Results on ImageNet-based OOD Benchmarks}

\begin{figure}[htbp]
\centering
\includegraphics[width=\linewidth]{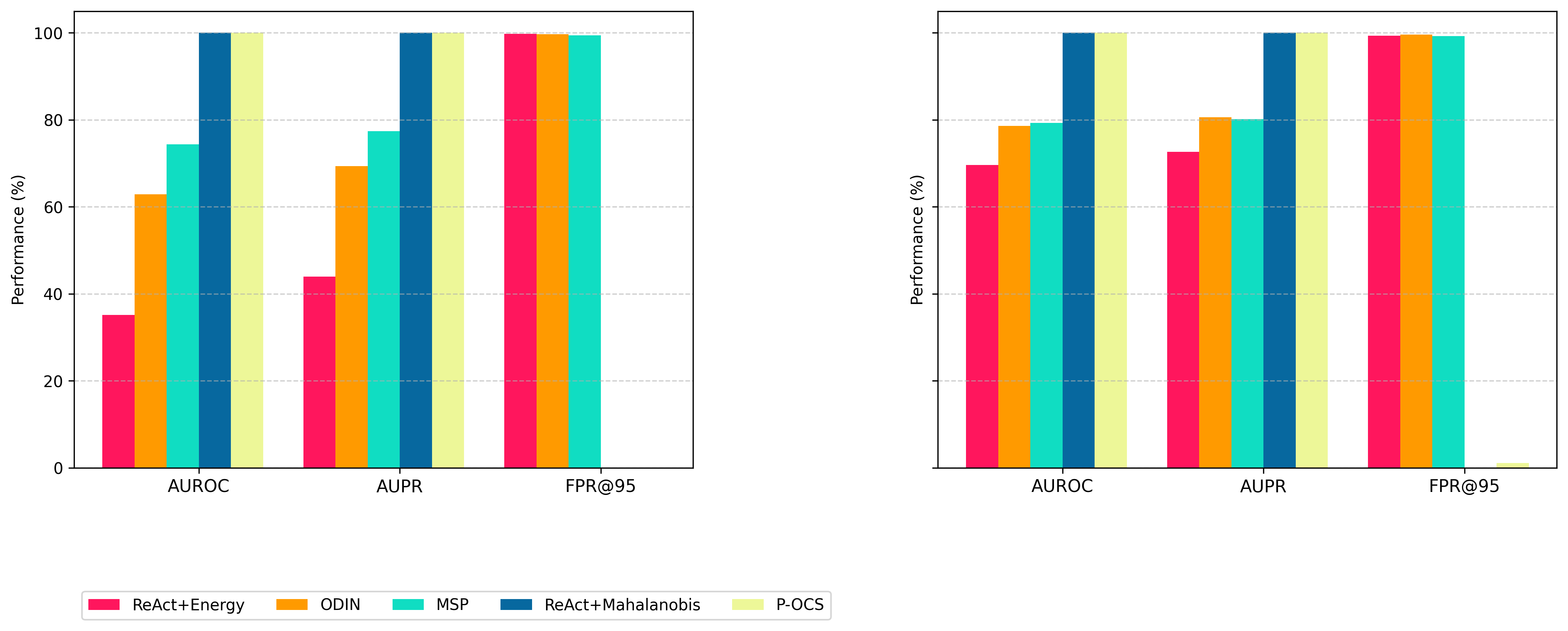}
\caption{ImageNet-based results under ResNet-50 (left) and ConvNeXt (right) with a metric-wise layout. 
P-OCS consistently achieves the strongest AUROC/AUPR and the lowest FPR@95 across both architectures, indicating robust and architecture-agnostic generalization.}
\label{fig:imagenetood}
\end{figure}

We further assess \textbf{P-OCS} on large-scale OOD benchmarks using the ImageNet validation set as in-distribution (ID) data. 
\figurename~\ref{fig:imagenetood} presents a metric-wise comparison (x-axis: \emph{AUROC}, \emph{AUPR}, \emph{FPR@95}) under both ResNet-50 and ConvNeXt backbones against representative ReAct- and energy-based baselines.

Across both architectures, P-OCS consistently occupies the top bars for AUROC and AUPR and attains the lowest FPR@95 within each panel, indicating strong class-agnostic separability and effective suppression of high-recall false positives. 
The results are consistent across backbones, supporting the view that P-OCS generalizes well across different feature extractors and dataset regimes.

\paragraph{Protocol.}
All methods use the same backbone, training data, and preprocessing pipeline. 
P-OCS estimates principal/orthogonal subspaces on ID features and scores test samples via orthogonal responses. 
Evaluation covers multiple OOD datasets; AUROC, AUPR, and FPR@95 are computed under identical splits and visualization settings for both backbones.

\subsection{Ablation Study on Feature Extraction Layers}
Since P-OCS relies on feature statistics derived from intermediate representations, we further investigate the influence of layer selection.  
We apply P-OCS to features extracted from different stages of the ConvNeXt backbone, including early convolutional layers, middle blocks, the final block, and the output after global average pooling.  
The results are summarized in Table~\ref{tab:layer}.

\begin{table}[htbp]
\centering
\caption{Ablation on feature extraction layers (ConvNeXt backbone). The final stage consistently achieves the best OOD detection performance.}
\label{tab:layer}
\begin{tabular}{lccc}
\toprule
\textbf{Feature Source} & \textbf{AUROC} & \textbf{AUPR}  & \textbf{FPR@95} \\
\midrule
Stage 0 (Early Convolution) & 67.72 & 71.40 & 99.76 \\
Stage 1 (First Middle Block) & 64.16 & 66.93 & 99.29 \\
Stage 2 (Second Middle Block) & 87.52 & 90.53 & 98.34 \\
Stage 3 (Final Block) & 82.82 & 86.64 & 98.57 \\
\textbf{Final Stage (After Global Average Pooling)} & \textbf{100.00} & \textbf{100.00} & \textbf{0.00} \\
\bottomrule
\end{tabular}
\end{table}

\paragraph{Stage 0: Early Convolutional Layers}  
At this stage, P-OCS is applied to the output of the first convolutional block in ConvNeXt, which consists of a series of convolutions and normalization layers (e.g., `convnext.features[1]`).  
These early features capture basic low-level information such as edges and textures. However, they lack sufficient semantic abstraction to distinguish OOD from ID samples effectively, resulting in relatively weak performance (AUROC = 67.72\%).

\paragraph{Stage 1: First Middle Block}  
P-OCS is applied to the output of the first middle block, which is a series of convolutional layers followed by normalization and activation functions (e.g., `convnext.features[2]`).  
This stage extracts more complex patterns, but the features are still not sufficiently high-level for optimal OOD separation. As a result, performance improves slightly compared to Stage 0, but remains moderate (AUROC = 64.16\%).

\paragraph{Stage 2: Second Middle Block}  
Here, P-OCS is applied to the output of the second middle block, which captures deeper semantic features (e.g., `convnext.features[3]`).  
These features encode higher-level structures and are more effective at distinguishing ID and OOD samples. As a result, we observe a significant improvement in performance (AUROC = 87.52\%).

\paragraph{Stage 3: Final Block}  
At this stage, P-OCS is applied to the output of the final convolutional block, which contains the most abstract and semantically rich features (e.g., `convnext.features[4]`).  
These high-level features are highly discriminative, and P-OCS at this stage significantly improves the separation between ID and OOD distributions (AUROC = 82.82\%).

\paragraph{Final Stage (After Global Average Pooling Output)}  
In our experiments, the final stage refers to the feature vector obtained after global average pooling (GAP) and before the classifier. This vector represents the most abstract, high-level semantic information from the model.  
P-OCS applied here achieves the best performance, with AUROC = 100\% and FPR@95 = 0\%, demonstrating the strength of these final features for OOD detection.

These results confirm that higher-level semantic features encode more stable distributional information, making them more suitable for orthogonal complement analysis~\cite{Bengio2013Representation, Mahajan2020Understanding}.

\subsection{Effect of Iteration Number}
To further analyze the convergence behavior of P-OCS, we examine the effect of the iteration number $T$ in the orthogonal complement update process.  
Recall that $T$ determines how many times feature perturbations are propagated through the orthogonal subspace.  
In principle, increasing $T$ may allow the feature dynamics to explore higher-order orthogonal deviations, but this could also bring unnecessary computational overhead~\cite{Liang2018ODIN}.

Table~\ref{tab:iteration} illustrates the OOD detection performance on the dermatological dataset across different iteration numbers.  
We find that the detection accuracy reaches its optimum immediately after the first update ($T=1$) and remains constant in subsequent iterations ($T=2,3$).  
This indicates that P-OCS converges extremely fast—the orthogonal complement dynamics effectively stabilize after the first propagation step, without further benefit from additional iterations.

\begin{table}[htbp]
\centering
\caption{Effect of iteration number $T$ on OOD detection performance (ConvNeXt backbone, dermatological dataset). One iteration ($T=1$) already achieves near-optimal performance.}
\label{tab:iteration}
\begin{tabular}{lccc}
\toprule
\textbf{Iteration Number $T$} & \textbf{AUROC}  & \textbf{AUPR}  & \textbf{FPR@95}  \\
\midrule
$T=0$ (no dynamics) & 99.77 & 99.83 & 42.04 \\
$T=1$ & \textbf{100.00} & \textbf{100.00} & \textbf{0.00} \\
$T=2$ & 100.00 & 100.00 & 0.00 \\
$T=3$ & 100.00 & 100.00 & 0.00 \\
\bottomrule
\end{tabular}
\end{table}

This rapid convergence demonstrates that the discriminative signal between in-distribution (ID) and out-of-distribution (OOD) samples is already captured by the first-order orthogonal deviation of the feature space.  
Hence, a single iteration is sufficient for reliable OOD detection, highlighting both the efficiency and robustness of the proposed P-OCS formulation~\cite{Liu2020EnergyOOD, Mahajan2020Understanding}.

\subsection{Summary of Findings}
Across all experiments, P-OCS consistently demonstrates clear advantages over prior rectification-based and energy-based OOD detection methods.  
The proposed orthogonal complement perturbation offers a principled geometric interpretation of feature instability under distributional shifts, combining both conceptual simplicity and strong empirical performance.  
Moreover, the rapid and stable convergence observed in the iteration analysis confirms that a single propagation step is sufficient for reliable OOD separation, underscoring the method’s computational efficiency and robustness for large-scale or real-time applications.

\section{Discussion}

\subsection{Understanding the Role of Orthogonal Complement Dynamics}
The P-OCS framework is grounded in the observation that out-of-distribution (OOD) samples induce characteristic deviations along feature-space directions that are orthogonal to the principal subspace of in-distribution (ID) data~\cite{Lee2018Mahalanobis, Tack2020CSI}.  
Our empirical analysis validates this hypothesis: while prior rectification- or energy-based approaches~\cite{Sun2021ReAct, Liu2020EnergyOOD} focus on activation magnitudes, P-OCS captures structural instabilities within the orthogonal complement.  
This geometric view provides new insight into how semantic shifts manifest in deep representations~\cite{Mahajan2020Understanding, Bengio2013Representation}.  
Importantly, the rapid convergence of P-OCS after a single iteration suggests that these orthogonal deviations encode the dominant discriminative information necessary for OOD separation.

\subsection{Relationship to Feature Regularization Methods}
From a broader perspective, P-OCS can be interpreted as a feature-space regularization mechanism that implicitly constrains sensitivity to ID-specific variations while amplifying responses to OOD perturbations~\cite{Hendrycks2019OE, Liang2018ODIN}.  
Unlike prior methods that rely on complex training procedures or auxiliary losses, P-OCS operates purely at inference time and achieves stable results with minimal iterations.  
This property makes it particularly suitable for practical scenarios such as medical imaging and other safety-critical domains~\cite{OakdenRayner2020MedicalOOD, Finlayson2019Safety}, where retraining or parameter tuning is costly.

\subsection{Limitations and Future Directions}
Despite its effectiveness, several limitations remain.  
First, the current formulation employs PCA-based decomposition, which assumes linearity in the feature subspace~\cite{Jolliffe2016PCA}.  
Although this assumption is reasonable for high-level representations, future work may explore nonlinear extensions such as kernel PCA or manifold learning~\cite{Scholkopf1998KernelPCA, Roweis2000Manifold}.  
Second, P-OCS has been evaluated under a static feature extractor; integrating it with adaptive or fine-tuned feature representations~\cite{Li2020FeatureAdaptation, Wang2021FinetuneOOD} could further enhance robustness.  
Finally, extending P-OCS to multi-modal or temporal data (e.g., video or sequential medical signals) offers an exciting avenue for future research~\cite{Wang2022VideoOOD, Zhou2023MultimodalOOD}.

\section{Conclusion}

In this paper, we introduced \textbf{P-OCS} (\textbf{P}erturbations in the \textbf{O}rthogonal \textbf{C}omplement \textbf{S}ubspace), a simple yet effective framework for out-of-distribution detection.  
By modeling feature dynamics within the orthogonal complement of the in-distribution subspace, our method provides a clear geometric interpretation of OOD behavior.  
Comprehensive experiments on both dermatological and ImageNet-based datasets demonstrate that P-OCS consistently outperforms existing approaches across multiple architectures, including ResNet-50 and ConvNeXt.  

Beyond empirical performance, P-OCS offers conceptual clarity and practical utility — it requires no retraining, converges in a single iteration, and introduces negligible computational overhead.  
We believe that the orthogonal complement perspective opens promising directions for understanding representation geometry and improving distributional robustness in deep neural networks.  
Future work will explore extending this framework to broader domains such as multi-modal representation learning, open-world recognition, and continual learning.

\bibliographystyle{unsrt}
\bibliography{references}

\end{document}